\documentclass{article}

\usepackage{arxiv}
\usepackage{graphicx}
\usepackage{natbib}

\usepackage[utf8]{inputenc} 
\usepackage[T1]{fontenc}    
\usepackage{hyperref}       
\usepackage{url}            
\usepackage{booktabs}       
\usepackage{amsfonts}       
\usepackage{nicefrac}       
\usepackage{microtype}      
\usepackage{lipsum}

\title{Modelling response to trypophobia trigger using intermediate layers of ImageNet networks}

\author{Piotr Woźnicki, Michał Kuźba, Piotr Migdał}

\begin{document}
\maketitle

\begin{abstract}
In this paper, we approach the problem of detecting trypophobia triggers using Convolutional neural networks. We show that standard architectures such as VGG or ResNet are capable of recognizing trypophobia patterns. We also conduct experiments to analyze the nature of this phenomenon. To do that, we dissect the network decreasing the number of its layers and parameters. We prove, that even significantly reduced networks have accuracy above 91\% and focus their attention on the trypophobia patterns as presented on the visual explanations.
\end{abstract}

\keywords{Trypophobia \and Computer vision \and Neuroscience \and Visual explanation \and Cognitive science}

\section{Introduction} \par
Trypophobia is a condition where a person experiences an intense and disproportionate fear towards clusters of small holes. The triggering images involve in general patterns of high-contrast energy at low and midrange spatial frequencies (\cite{martinez}). It has been suggested in the literature that aversion to clusters is an evolutionarily prepared response to stimuli that resemble parasites and infectious disease (\cite{cole}). This generally adaptive phenomenon may become maladaptive if a strong aversive response is caused by harmless triggers. \\ 
In this paper, we are modelling trypophobia response using Convolutional Neural Networks. These models use the idea of local receptive fields, which has been originally inspired by mammalian visual system (\cite{lecun-gradientbased-learning-applied-1998}). First, we check how accurate different CNN architectures are in classifying an image as trypophobic. \\ 
It has been shown that textures and patterns are interpreted by early layers of a CNN and are further being combined in the following layers to build up the understanding of complicated objects and relationships (\cite{olah2017feature}). Therefore, we make further experiments with reducing the number of convolution blocks and reducing the network processing pipeline in order to see how early could the trypophobia pattern be recognized. It is known that threatening cues in the environment receive priority in visual attention processing and automatically capture attention (\cite{Shirai2019TrypophobicII}). We state the hypothesis that trypophobia cues can be effectively recognized by a simplified  neural network with massively reduced number of layers and parameters. 

\section{Dataset and preprocessing}
The dataset we used had been prepared by \cite{puzio19}. Our training data consists of 6000 trypophobia triggering  and 10000 neutral images. The test set contains 500 trypophobic and 500 neutral images. For training, we use standard data augmentation techniques including horizontal/vertical mirroring, rotation in range of 45$^\circ$, shear and zoom in range 0.3.

\section{Experiment design}
We train a few different models based on popular CNN architectures, including VGG16, Inceptionv3 and ResNet50 for the task of classifying image as trypophobic/neutral. We also experiment with using pretrained weights from ImageNet and freezing convolutional blocks.

\begin{figure}[hb]
  \includegraphics[width=.8\linewidth]{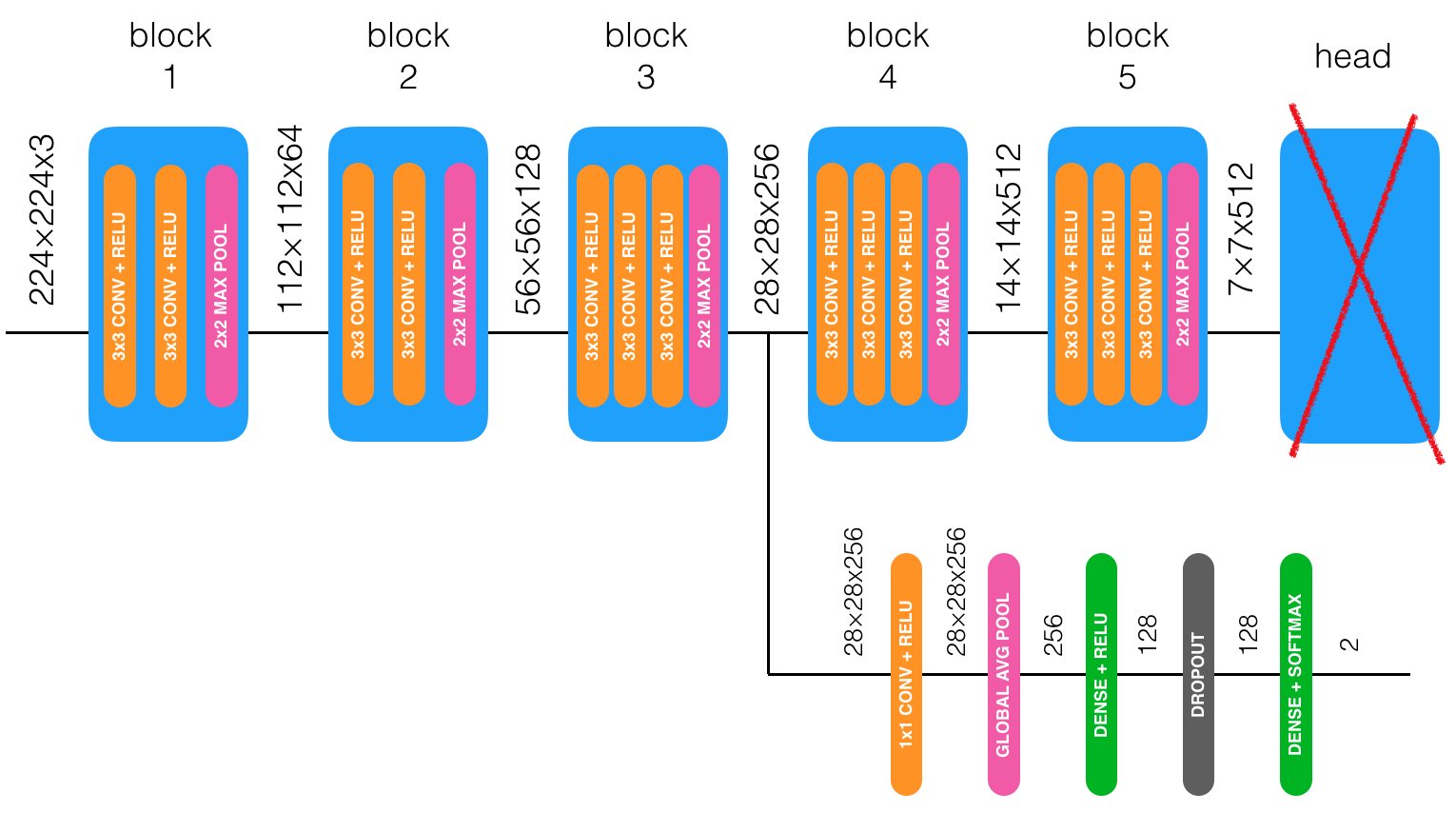}
  \centering
  \caption{\footnotesize Model architecture: after $k$ convolutional blocks of ImageNet-pretrained VGG16, we use a fully convolutional head. We use only 1x1 convolution, to avoid introducing new spatial features. The size of hidden layer (here: $128$) and the dropout rate (here: $0.5$) are our hyperparameters.}
  \label{architecture}
\end{figure}

For further experiments we choose the best performing VGG16 model (as proposed by \cite{simonyan2014deep}), because of its relative simplicity and as good performance as more complicated networks. We create four new models by using only respectively the first 4, 3, 2 and 1 Conv blocks, as shown on Figure~\ref{architecture}. We add a 1x1 Conv layer after the last Conv block and experiment with the the setup of top layers, choosing between Global average and Max pooling, using 1 or more Dense layers before the final classification and adjusting the size of Dense layers (between 32 and 1024 nodes). We set a constant number of 50 epochs and tune the learning rate for each model separately. For networks with smaller number of Conv blocks we experiment with downsampling the original image by a factor of 2 and 4 to make up for bigger feature maps before Global pooling layer. We use Dropout regularization for the dense layers. The results are presented in the next section. Our code can be found on github: \url{https://github.com/kmichael08/trypophobia-detection}.

\begin{table}[t]
\footnotesize
\setlength{\tabcolsep}{2 pt}
\centering
\caption{\footnotesize Results for the binary classification for 'cut' VGG16 models.}

\begin{tabular}{l||c|c|c|c|c|c|c}
\textbf{Architecture} & \textbf{Full model} & \textbf{4 Conv blocks} & \textbf{3 Conv blocks} & \textbf{2 Conv blocks} & \textbf{1 Conv block} \\ \hline
Test accuracy & \textbf{94.5} & \textbf{94.0} & \textbf{92.9} & \textbf{91.0} & \textbf{82.8}\\ 
Test AUC & 0.988 & 0.987 & 0.981 & 0.974 & 0.921 \\
Train accuracy & 98.2 & 96.1 & 96.1 & 92.6 & 84.7 \\ \hline
Num of parameters &  \textbf{17 926 209} &  \textbf{8 161 089} &  \textbf{1 867 329} &  \textbf{293 313} &  \textbf{47 105} \\ \hline
\end{tabular}
\label{test}

\end{table}

\section{Results}
The architectures from the first experiment achieve 90.0-94.5\% accuracy, depending on a specific configuration. We observed that both using transfer learning in the form of \textit{ImageNet} weights as well as unfreezing the conv layers are beneficial. \\
Next, we use the VGG16 to compare it with its 'cut' versions, as described in the previous section. We evaluate them on the test set. The best-performing design is shown below and the results for 'cut' models with this design are presented in Table~\ref{test}.

\begin{figure}[ht]
    \includegraphics[width=0.85\linewidth]{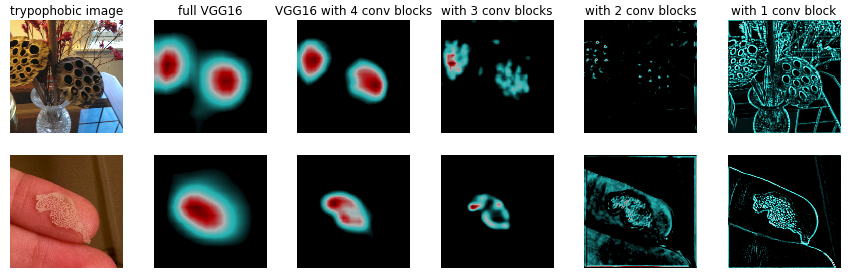}
\centering
\caption{\footnotesize Gradient-weighted Class Activation Mapping for 2 trypophobic images.}
\label{gradcam}

\end{figure}

We report accuracy ranging from 82.8\% to 94.0\%, as compared to 94.5\% achieved by the full VGG16 model. In our experiments we have consistently seen networks with 4 and 3 Conv blocks performing comparably to the full VGG16 and being more stable while training and less prone to overfitting. We have not observed improvement for downsampled images, but further experiments should be performed to see if we could retain good performance with further reduction of number of parameters. \\
We apply Gradient-weighted Class Activation Mapping (Grad-CAM)(\cite{DBLP:journals/corr/SelvarajuDVCPB16})  method to check visual explanations for detecting trypophobia. Two correctly classified trypophobic images are shown as an example in Figure~\ref{gradcam}. It seems that even the networks with 1 and 2 Conv blocks focus their attention on the trypophobia pattern in the image.

\section{Conclusions}
CNNs achieve $>94\%$ accuracy on detecting trypophobia triggers. However, shallower architectures can be effectively trained to achieve performances matching these of deep models. We showed that having reduced the number of parameters of the model by $98.4\%$ to just 2 convolutional blocks instead of 5, we still retain a high accuracy of $91\%$. It means that a CNN understands very early, that a triggering pattern is present. We suggest that patterns like trypophobia are of a primitive nature and can be quickly and effectively detected as such by human visual system. These threat-related stimuli are given high priority in the brain to automatically capture attention and subsequently influence behavior.

\bibliographystyle{abbrvnat}
\bibliography{references}

\begin{thebibliography}{8}
\providecommand{\natexlab}[1]{#1}
\providecommand{\url}[1]{\texttt{#1}}
\expandafter\ifx\csname urlstyle\endcsname\relax
  \providecommand{\doi}[1]{doi: #1}\else
  \providecommand{\doi}{doi: \begingroup \urlstyle{rm}\Url}\fi

\bibitem[Cole and Wilkins(2013)]{cole}
G.~Cole and A.~Wilkins.
\newblock Fear of holes.
\newblock \emph{Psychological science}, 24, 08 2013.
\newblock \doi{10.1177/0956797613484937}.

\bibitem[LeCun et~al.(1998)LeCun, Bottou, Bengio, and
  Haffner]{lecun-gradientbased-learning-applied-1998}
Y.~LeCun, L.~Bottou, Y.~Bengio, and P.~Haffner.
\newblock Gradient-based learning applied to document recognition.
\newblock In \emph{Proceedings of the IEEE}, volume~86, pages 2278--2324, 1998.
\newblock URL
  \url{http://citeseerx.ist.psu.edu/viewdoc/summary?doi=10.1.1.42.7665}.

\bibitem[Martínez et~al.(2018)Martínez, Lanfranco, Arancibia, Sepúlveda, and
  Madrid~Aris]{martinez}
J.~Martínez, R.~Lanfranco, M.~Arancibia, E.~Sepúlveda, and E.~Madrid~Aris.
\newblock Trypophobia: What do we know so far? a case report and comprehensive
  review of the literature.
\newblock \emph{Frontiers in Psychiatry}, 9, 02 2018.

\bibitem[Olah et~al.(2017)Olah, Mordvintsev, and Schubert]{olah2017feature}
C.~Olah, A.~Mordvintsev, and L.~Schubert.
\newblock Feature visualization.
\newblock \emph{Distill}, 2017.
\newblock \doi{10.23915/distill.00007}.
\newblock https://distill.pub/2017/feature-visualization.

\bibitem[Puzio and Uriasz(2019)]{puzio19}
A.~Puzio and G.~Uriasz.
\newblock Trypophobia images detector based on deep neural networks and
  utilities.
\newblock \url{https://github.com/cytadela8/trypophobia}, 2019.

\bibitem[Selvaraju et~al.(2016)Selvaraju, Das, Vedantam, Cogswell, Parikh, and
  Batra]{DBLP:journals/corr/SelvarajuDVCPB16}
R.~R. Selvaraju, A.~Das, R.~Vedantam, M.~Cogswell, D.~Parikh, and D.~Batra.
\newblock Grad-cam: Why did you say that? visual explanations from deep
  networks via gradient-based localization.
\newblock \emph{CoRR}, abs/1610.02391, 2016.
\newblock URL \url{http://arxiv.org/abs/1610.02391}.

\bibitem[Shirai et~al.(2019)Shirai, Banno, and Ogawa]{Shirai2019TrypophobicII}
R.~Shirai, H.~Banno, and H.~Ogawa.
\newblock Trypophobic images induce oculomotor capture and inhibition.
\newblock \emph{Attention, Perception, \& Psychophysics}, 81:\penalty0
  420--432, 2019.

\bibitem[Simonyan and Zisserman(2014)]{simonyan2014deep}
K.~Simonyan and A.~Zisserman.
\newblock Very deep convolutional networks for large-scale image recognition,
  2014.

\end{thebibliography}

\end{document}